\def\year{2019}\relax
\documentclass[letterpaper]{article} 
\usepackage{aaai19}  
\usepackage{times}  
\usepackage{helvet}  
\usepackage{courier}  
\usepackage{url}  
\usepackage{graphicx}  
\usepackage{color}
\usepackage[british]{babel}
\usepackage{amsfonts}
\usepackage{amsmath}
\usepackage{dsfont}
\usepackage{mathtools, xparse}
\usepackage{algorithm}
\usepackage{algorithmic}
\usepackage{multirow}
\usepackage[normalem]{ulem}
\usepackage{booktabs}
\usepackage{subcaption}

\begin{document}
 
\newcommand{\eref}[1]{(\ref{#1})}
\newcommand{\sref}[1]{Section~\ref{#1}}
\newcommand{\figref}[1]{Fig.~\ref{#1}}
\newcommand{\tabref}[1]{Table~\ref{#1}}
\newcommand{\appref}[1]{Appendix \ref{#1}}

\newcommand{\citep}[1]{\citeauthor{#1} \citeyear{#1}}
\newcommand{\R}{\mathcal{R}}

\newcommand{\ssnote}[1]{{\xxnote{SS}{red}{#1}}}
\newcommand{\lpnote}[1]{{\xxnote{LP}{blue}{#1}}}
\newcommand{\avknote}[1]{{\xxnote{AVK}{cyan}{#1}}}
\newcommand{\bhnote}[1]{{\xxnote{BH}{magenta}{#1}}}
\newcommand{\xxnote}[3]{}
\ifx\hidenotes\undefined
  \renewcommand{\xxnote}[3]{\color{#2}{#1: #3}}
\fi

\newcommand{\planner}{\text{PC-RRT}}
\newcommand{\our}{\text{LeaPER}}

%
\title{Sample-Efficient Learning of Nonprehensile Manipulation Policies via \\ Physics-Based Informed State Distributions}


\author{Lerrel Pinto \\ Carnegie Mellon University \\ \tt{lerrelp@cs.cmu.edu} \And Aditya Mandalika \\ University of Washington \\ \tt{adityavk@cs.uw.edu} \And Brian Hou \\ University of Washington \\ \tt{bhou@cs.uw.edu}\And Siddhartha Srinivasa \\ University of Washington \\ \tt{siddh@cs.uw.edu}}
\maketitle
\frenchspacing
\begin{abstract}
This paper proposes a sample-efficient yet simple approach to learning closed-loop policies for nonprehensile manipulation. Although reinforcement learning (RL) can learn closed-loop policies without requiring access to underlying physics models, it suffers from poor sample complexity on challenging tasks. 
To overcome this problem, we leverage rearrangement planning to provide an informative physics-based prior on the environment's optimal state-visitation distribution. Specifically, we present a new technique, Learning with Planned Episodic Resets ($\our$), that resets the environment's state to one informed by the prior during the learning phase. We experimentally show that $\our$ significantly outperforms traditional RL approaches by a factor of up to 5X on simulated rearrangement. Further, we relax dynamics from quasi-static to welded contacts to illustrate that LeaPER is robust to the use of simpler physics models. Finally, LeaPER's closed-loop policies significantly improve task success rates relative to both open-loop controls with a planned path or simple feedback controllers that track open-loop trajectories. We demonstrate the performance and behavior of $\our$ on a physical 7-DOF manipulator in \url{https://youtu.be/feS-zFq6J1c}.


\end{abstract}
\section{Introduction}

\begin{figure}[ht]
\centering
\includegraphics[width=3.3in]{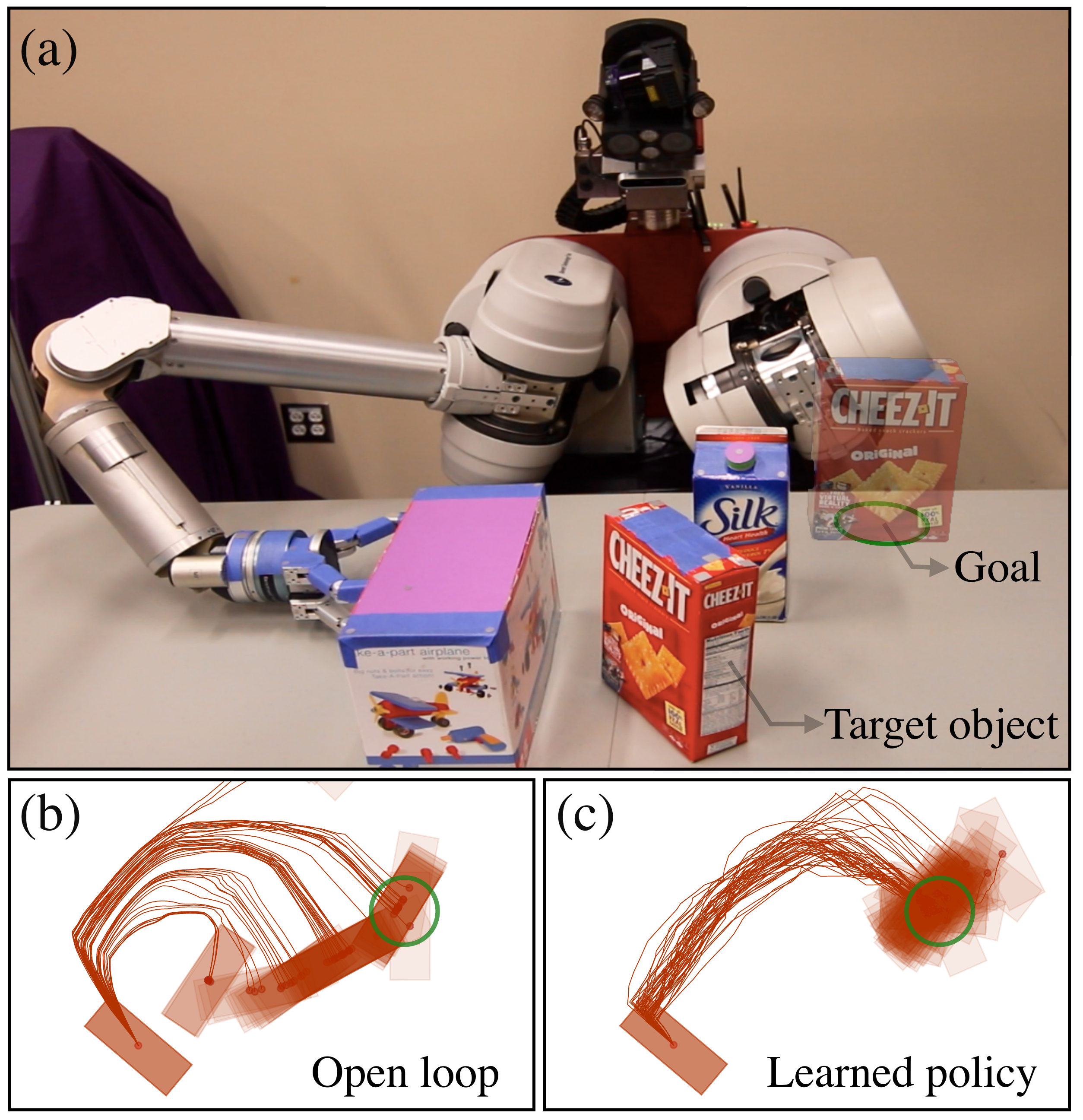}
\caption{
(a) The goal of rearrangement manipulation is to move a target object to a desired goal region (green).
(b) Object trajectories diverge when running open-loop controls from the planner $\planner$.
(c) Trajectories converge when using $\our$ policies.}
\label{fig:intro}
\end{figure}

\begin{figure*}[ht]
\centering
\includegraphics[width=7.0in]{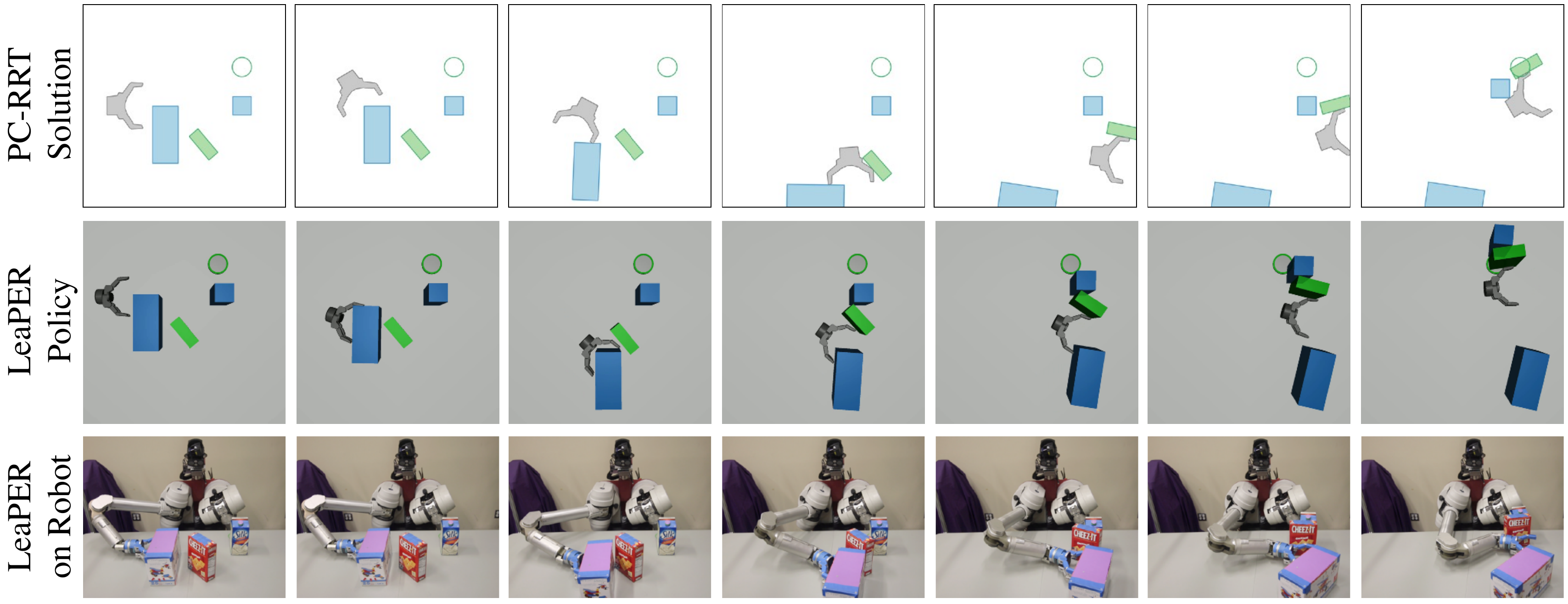}
\caption{We use the $\planner$ planner to generate a fast open-loop solution (top row) for the rearrangement problem with quasi-static physics. Using this solution as a prior, $\our$ learns a policy using full-contact physics simulations (middle row), which is then executed on a 7-DOF robot manipulator (bottom row).}
\label{fig:desc}
\end{figure*}

This paper addresses the problem of learning closed-loop policies for nonprehensile rearrangement. Consider the environment in \figref{fig:intro}a. The robot must push the target object from the front of the table to the desired goal at the back of the table. Rearrangement is frequently needed for robots to manipulate diverse objects in cluttered environments.  To perform this task successfully, the robot must reach and carefully manipulate the box through a clutter of movable objects. Since this involves making and breaking contact with multiple objects over long horizons, rearrangement manipulation poses a challenging robotic task. 

Previous work has considered rearrangement to be a planning problem, where a planner searches for feasible paths that connect the initial configuration of the environment to the desired goal configuration. The solution is then executed in an open-loop manner to perform the rearrangement task. These methods successfully generate expressive solutions~\cite{king2016rearrangement}. However, a key component of these planning algorithms is that they employ approximate but fast contact models to roll out simulations~\cite{lynch1996stable}. Hence, the open-loop paths that the planner generates are only realizable in the real world if the approximate simulations sufficiently mimic the real world. Due to the complexities of real contact physics, contact dynamics are often significantly relaxed for efficient planning. This causes the generated open-loop trajectories to diverge from the desired behavior and fail to reach the goal configuration (\figref{fig:intro}b). 

An alternative approach is to use deep reinforcement learning (RL) to learn closed-loop policies. These policies take observations of the environment as inputs and output actions that move the target object to its desired goal. Since they observe and act at a high frequency, the policies can correct for deviations caused by uncertainty in dynamics and other modelling errors. However, RL has notoriously high sample complexity~\cite{duan2016benchmarking}, which has limited its use to single object interactions~\cite{agrawal2016learning} or with carefully designed dense reward functions~\cite{brockman2016openai}. Interacting with multiple objects is fundamental to rearrangement, which also makes designing reward functions impractical.

The high sample complexity of RL can be attributed to the trade-off between exploration and exploitation~\cite{kearns2002near}, i.e. should an agent exploit its current experience to choose the seemingly best action or should it explore the environment by taking different actions that may lead to better future payouts? If the agent does not sufficiently explore, it may not converge to the optimal behavior. However, if the agent explores too much, it will require more episodes before converging on the optimal behavior. 
Considerable research has focused on exploration techniques for RL agents. One powerful technique is to force the policy to visit certain relevant states. CPI~\cite{kakade2002approximately} sets the initial state distribution to a uniform distribution over the entire state space, which asymptotically ensures visiting every possible state. Although this approach provides strong guarantees of convergence, it is not feasible for the large state-space of rearrangement manipulation. A more informed distribution can be obtained from human demonstrations; ~\citep{nair2017overcoming}, and ~\citep{hosu2016playing} demonstrated significant improvements for manipulation tasks and Atari games. But how can we obtain these priors without expert demonstrations?



We contribute three key insights to more efficiently solving the rearrangement problem. Our first key insight is that the open-loop trajectories generated via rearrangement planning provide a powerful source of relevant states (\figref{fig:desc}). Although the quasi-static physics models used while planning are approximate, we show that the states visited by a solution are informative and significantly reduce our algorithm's sample complexity. Experimentally, {Learning with Planned Episodic Resets} ($\our$) accelerates learning by up to 5X on simulated rearrangement tasks. 


Our second insight is that planning with even simpler physics models can generate solutions that are sufficiently informative to accelerate RL for rearrangement. To illustrate this, we further relax the dynamics from quasi-static contacts to welded contacts. Although this simplification increases sample complexity, it is significantly superior in performance to not using planned resets. More importantly, using simplified contact models makes the planning problem much easier to solve.

Our third insight is that $\our$'s closed-loop policies are robust to stochastic dynamics since they can correct for deviations from the nominal path. Using $\our$-learned closed-loop policies, we observe significant improvements in task success ratios compared to open-loop controls with a planned path or simple feedback controllers that track open-loop trajectories. We apply these robust policies, to demonstrate $\our$'s ability to solve the rearrangement manipulation task with a 7-DOF manipulator (\figref{fig:desc}).

\section{Background}

Before describing our work, we briefly introduce prerequisite background information and a formalism for rearrangement planning and reinforcement learning. We refer the reader to ~\citep{latombe2012robot} and ~\citep{kaelbling1996reinforcement} for a more comprehensive introduction to these topics.

In rearrangement manipulation, a robot $\mathcal{R}$ is tasked with moving $m$ movable objects $\mathcal{M}$ to desired configurations in the workspace.
However, there are obstacles $\mathcal{O}$ that the robot is forbidden to penetrate.

\subsection{Solving the Rearrangement Planning Problem}

Rearrangement manipulation can be considered a planning problem in a high-dimensional space that captures all possible configurations of the robot and other movable objects in the environment. The state space $C$ in which we plan is therefore a Cartesian product space of the robot's configuration space $C^\mathcal{R}$ and the configuration spaces $C^i,~\forall i \in 1,\ldots,m$ corresponding to each movable object, i.e., $C = C^\mathcal{R} \times C^1 \times \ldots \times C^m$. Each state $q \in C$ is defined as $q = \left(q^\mathcal{R}, q^1, q^2, \ldots, q^m\right)$, where $q^\mathcal{R} \in C^\mathcal{R},~q^i \in C^i$. The free state space $C_{\textrm{free}} \subseteq C$ is the set of all states where the robot and objects do not penetrate into each other or into obstacles. Contact between the physical entities is allowed, since it is necessary for rearrangement manipulation.

The rearrangement manipulation planning problem is to find a \emph{feasible} trajectory $\xi : \mathbb{R}^{\geq 0} \rightarrow C_{\textrm{free}}$ starting from  $\xi(0)\in C_{\textrm{free}}$ and ending in the goal region $\xi(T) \in C_G \subseteq C_{\textrm{free}}$ at some time $T\geq 0$. Since the evolution of the state $q$ is governed by the non-linear physics of contact and interaction between the robot and the objects, the states along the trajectory must satisfy the non-holonomic constraint $\dot{q} = f(q,u)$. Here, $u\in \mathcal{U}$ is the instantaneous control input, and $f$ encodes the physics of the environment. A path $\xi$ is feasible if there exists a control, $u$, at every time $0\leq t \leq T$ that satisfies the physics constraint $f$.

To find a feasible trajectory, we use a variant of Rapidly-Exploring Random Trees (RRT) \cite{lavalle1998rapidly} shown to be effective for planning in high-dimensional spaces with non-holonomic constraints. Specifically, we use the \emph{Physics-Constrained} RRT ($\planner$) planner~\cite{king2015nonprehensile}, which embeds a physics model in the traditional RRT algorithm. We constrain plans and corresponding actions to a manifold that is parallel to the table's surface. Moreover, to extend the search tree during planning, we use the quasi-static physics model~\cite{lynch1996stable} to propagate the control actions for a given control duration.\footnote{We refer the reader to the Appendix for more details.} This model assumes that the pushing motions are slow enough to render inertial forces negligible. Thus, we assume that objects move only when the robot contacts and pushes them. They immediately come to rest when the robot no longer applies a force on them. The constraint and the model therefore allow us to plan trajectories in a lower-dimensional space where each state $q$ is a set of SE(2) poses for the robot's end-effector and each movable object~\cite{lavalle2006planning}.

\subsection{Reinforcement Learning for Rearrangement}
Rearrangement manipulation can also be framed as a continuous space Markov Decision Process (MDP) represented by the tuple $(\mathcal{S},\mathcal{U},\mathcal{P},r,\gamma, \mathbb{S}, \mathbb{G})$, where $\mathcal{S}$ is a set of continuous states; $\mathcal{U}$ is a set of continuous actions; $\mathcal{P}: \mathcal{S} \times \mathcal{U} \times \mathcal{S} \rightarrow \mathbb{R}$ is the transition probability function; $r: \mathcal{S} \times \mathcal{U} \rightarrow \mathbb{R}$ is the reward function; $\gamma$ is the discount factor; $\mathbb{S}$ is the initial state distribution; and $\mathbb{G}$ is the desired goal distribution. The MDP state $s$ is related to the planning configuration state $x$ as $s = [x, \dot{x}]$. In the remainder of this paper, we use \emph{state} to refer to the MDP state $s$ and \emph{C-state} for the planning configuration state $x$.

An episode for the agent begins by sampling $s_0$ from the initial state distribution $\mathbb{S}$.
At every timestep $t$, the agent takes an action $u_t=\pi(s_t)$ according to a policy $\pi:\mathcal{S} \rightarrow \mathcal{U}$ to reach states in the goal distribution $\mathbb{G}$. To achieve the goal, the agent receives a reward $r_t=r(s_t,u_t)$. The action $u_t$ results in a state transition to $s_{t+1}$, which is sampled according to probabilities $\mathcal{P}(s_{t+1} | s_t,u_t)$.
The agent's goal is to maximize the expected return $E_{\mathbb{S}}[R_0|\mathbb{S}]$, where the return is the discounted sum of the future rewards $R_t=\sum^{\infty}_{i=t}\gamma^{i-t}r_i$. The $Q$-function is defined as $Q^{\pi}(s_t,u_t)=E[R_t|s_t,u_t]$. 

A key design decision is how to structure the reward function $r$ so that the robot achieves the desired goal configuration. Since rearrangement requires manipulating multiple objects, hand-designing a dense reward is challenging. Furthermore, recent work~\cite{andrychowicz2017hindsight} shows that densifying rewards often leads to unintended behavior. Hence, we use a sparse reward:
\begin{equation*}
    r(s_t,u_t)= \mathds{1}[s_t \in \mathbb{G}]
\end{equation*}

\subsection{Deep Deterministic Policy Gradients (DDPG)}
DDPG~\cite{lillicrap2015continuous} is an \textit{actor-critic} RL algorithm that learns a deterministic continuous action policy. The algorithm maintains two neural networks: the policy (or \emph{actor}) $\pi_\theta:\mathcal{S} \rightarrow \mathcal{U}$ with neural network parameters $\theta$ and a $Q$-function approximator (or \emph{critic}) $Q_\phi^{\pi}:\mathcal{S} \times \mathcal{U} \rightarrow \mathbb{R}$ with neural network parameters $\phi$. 

During training, episodes are generated using a noisy version of the policy (called the \emph{behavior policy}), 
e.g., $\pi_b(s) = \pi(s) + \mathcal{N}(0,1)$.
A replay buffer stores the transition tuples $(s_t,a_t,r_t,s_{t+1})$ encountered during training~\cite{mnih2015human}. Training examples sampled from the replay buffer are used to optimize the critic. By minimizing the Bellman error loss $\mathcal{L}_c=(Q(s_t,a_t)-y_t)^2$, where $y_t=r_t + \gamma Q(s_{t+1},\pi(s_{t+1}))$, the critic is optimized to approximate the $Q$-function. The actor is optimized by minimizing $\mathcal{L}_a=-E_s[Q(s,\pi(s))]$. The gradient of $\mathcal{L}_a$ with respect to the actor parameters can be computed by backpropagating through the combined critic and actor networks. 



\begin{figure*}[!ht]
\begin{center}
\includegraphics[width=7.0in]{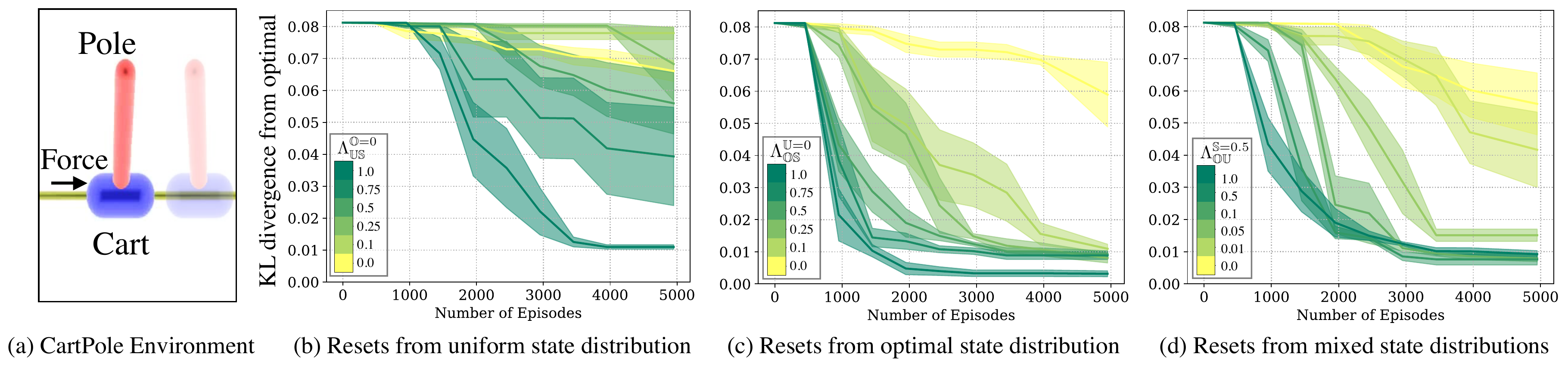}
\end{center}
\caption{
We illustrate
(a) the CartPole environment and show learning curves with varying 
(b) $\Lambda^{\mathbb{O}=0}_{\mathbb{U},\mathbb{S}}$, 
(c) $\Lambda^{\mathbb{U}=0}_{\mathbb{O},\mathbb{S}}$, and 
(d) $\Lambda^{\mathbb{S}=0.5}_{\mathbb{O},\mathbb{U}}$.
}
\label{fig:cartpole}
\end{figure*}

\subsection{Hindsight Experience Replay (HER)}

HER~\cite{andrychowicz2017hindsight} is a simple way to manipulate the replay buffer used in off-policy RL algorithms to learn policies more efficiently using sparse rewards. Instead of learning a policy that takes only $s_t$ as input, the policy also takes the desired goal as input, i.e., $u_t = \pi(s_t, g)$. Without HER, after experiencing some episode $s_0,s_1,...,s_T$, every transition $s_t\rightarrow s_{t+1}$ is stored with the episode's goal in the replay buffer. With HER, the replay buffer stores the experienced transitions but with different goals, i.e., states reached later in the episode. Since the goal being pursued does not influence the environment dynamics, we can replay each trajectory using arbitrary goals assuming we use an off-policy RL algorithm for optimization.
\section{Episodic Resets from Planned States}

\subsection{A motivating example}
Our method is centered around altering the initial state distribution during learning. To analyze its effect, we use the CartPole environment (\figref{fig:cartpole}a) as a controlled example. Given an initial state of the environment $s_0 \sim \mathbb{S}$, the agent must reach the goal $x_G$. The RL formulation is similar to rearrangement planning, where rewards are sparse and the agent has a fixed horizon. We examine three sources of initial states: (a) the test-time initial state distribution $\mathbb{S}$, (b) the uniform state distribution $\mathbb{U}$ \cite{kakade2002approximately} and (c) the optimal state visitation distribution $\mathbb{O}$. We compute the optimal policy $\pi^{*}$ using iLQR~\cite{li2004iterative} to generate $\mathbb{O}$.

We define $\Lambda^{A=a}_{B,C}$ as the ratio of sampling from $B$ and $C$, with the probability of sampling from $A$ fixed at $a$. \figref{fig:cartpole} illustrates the effect of changing the mixing distribution on the KL divergence between the resulting policy's state distribution and the optimal state distribution. Varying $\Lambda^{\mathbb{O}=0}_{\mathbb{U},\mathbb{S}}$ (\figref{fig:cartpole}b) shows how increasing samples from the uniform state distribution is better than sampling solely from $\mathbb{S}$. Varying $\Lambda^{\mathbb{U}=0}_{\mathbb{O},\mathbb{S}}$ (\figref{fig:cartpole}c) demonstrates how the learning rate can be further improved by sampling reset states from the optimal state visitation distribution $\mathbb{O}$. Finally, variations in $\Lambda^{\mathbb{S}=0.5}_{\mathbb{O},\mathbb{U}}$ (\figref{fig:cartpole}d) emphasize that even a small fraction of optimal states with uniformly sampled states can give a substantial speedup for learning. We note that although sampling from $\mathbb{U}$ is useful, sampling states uniformly may not capture the relevant state space for complex tasks. Sampling from the optimal state visitation distribution alleviates this issue and ensures that the sampled states are valid and collision-free (e.g. for rearrangement). With these insights, we now present $\our$.

\subsection{Learning from Planned Episodic Resets ($\our$)}

Our algorithm starts with the initial configuration of the environment $x_I$ and the desired goal space $X_G$. In practice, $X_G$ is defined by the $\epsilon$-ball around a single goal C-state $x_G$. The first part of our method needs a planner $P$ to solve the rearrangement task, which gives us a trajectory of states $\xi_P \equiv \{x_0, x_1, ..., x_T\}$. Here, $x_0 = x_I$ and $x_T \in X_G$. Note that although this trajectory is feasible with quasi-static physics, it may not be feasible with real physics. 

\begin{algorithm}
\begin{algorithmic}
\STATE \textbf{Given} Initial configuration $x_I$, desired goal space $X_G$
\STATE Initialize off-policy RL algorithm $\mathcal{A}$
\STATE Initialize policy $\pi^{0}$
\STATE Initialize planning algorithm $P$
\STATE Plan rearrangement trajectory $\xi_{P}$
\STATE \hspace{0.5in} {$\xi_{P} \leftarrow P(x_I,X_G)$}
\FOR{episode $m=1:M$} 
    \STATE Sample initial state $s_0$
    \STATE \hspace{0.5in} $s_0 \sim
    \begin{cases}
      \xi_{P}, & \alpha \\
      \mathbb{S}, & 1-\alpha
    \end{cases}$
    \STATE Collect episode $\tau^{m}$
    \STATE \hspace{0.5in} $\tau^{m} \leftarrow \text{rollouts}(s_0, \pi^{m})$
    \STATE Improve policy with RL algorithm $\mathcal{A}$
    \STATE \hspace{0.5in} $\pi^{m+1} \leftarrow \mathcal{A}(\tau^{m}, \pi^{m})$
\ENDFOR
\end{algorithmic}
\caption{$\our$}
\label{alg:ep_res}
\end{algorithm}

Given the planned trajectory $\xi_P$, we can now start the learning process. For each episode in the environment, the initial state $s_0$ is uniformly sampled from $\xi_P$ with probability $\alpha$. With probability $1-\alpha$, $s_0$ is sampled from the original start distribution. Given this initial state, an episode is collected by rolling out the policy $\pi^{m}$ in the physics environment. The RL algorithm $\mathcal{A}$ then optimizes the weights of $\pi$ and updates the policy to $\pi^{m+1}$. After $M$ episodes, the closed-loop policy rearranges objects from $x_I$ to $X_G$.

Since we have a sparse reward and a long horizon task, we use Hindsight Experience Replay (HER) with Deep Deterministic Policy Gradients (DDPG) as our base RL algorithm $\mathcal{A}$. For the planning algorithm $\mathcal{P}$, we use $\planner$ with quasi-static physics.
\section{Analyzing $\our$}

\begin{figure*}[!ht]
\begin{center}
\includegraphics[width=7.0in]{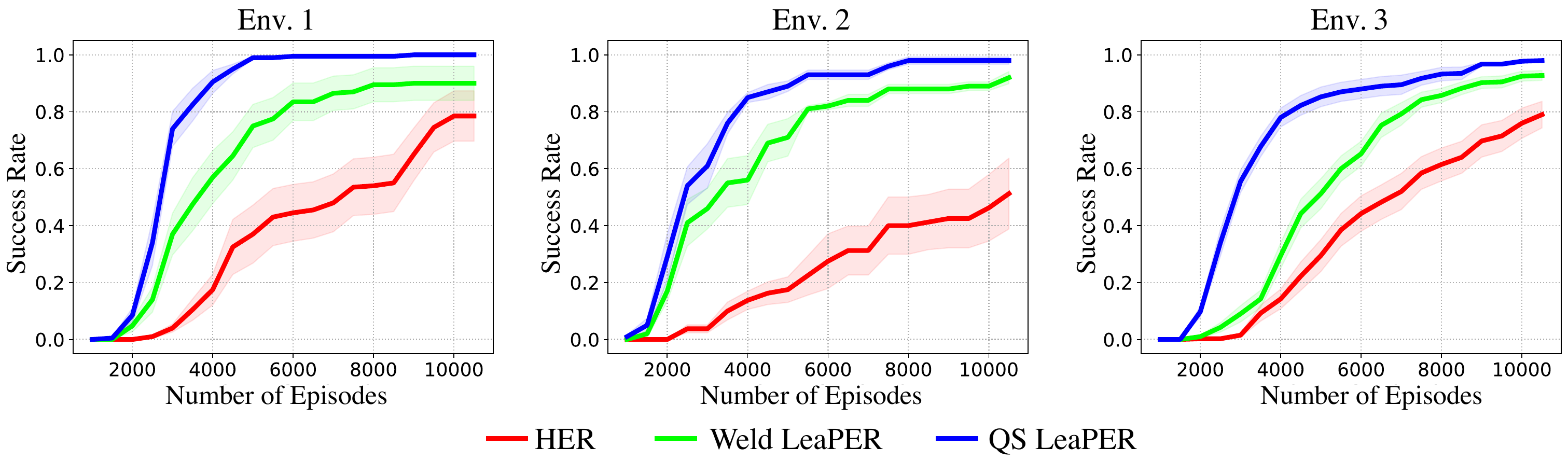}
\end{center}
\caption{The learning curves for HER, Weld $\our$ and quasi-static (QS) $\our$ are shown for Env. 1 (left), Env. 2 (middle), and Env. 3 (right). The shaded regions correspond to the 90\% confidence interval on the mean across 20 random seeds.}
\label{fig:curves}
\end{figure*}

\subsection{Environments}
\begin{figure}[]
\begin{center}
\includegraphics[width=\linewidth]{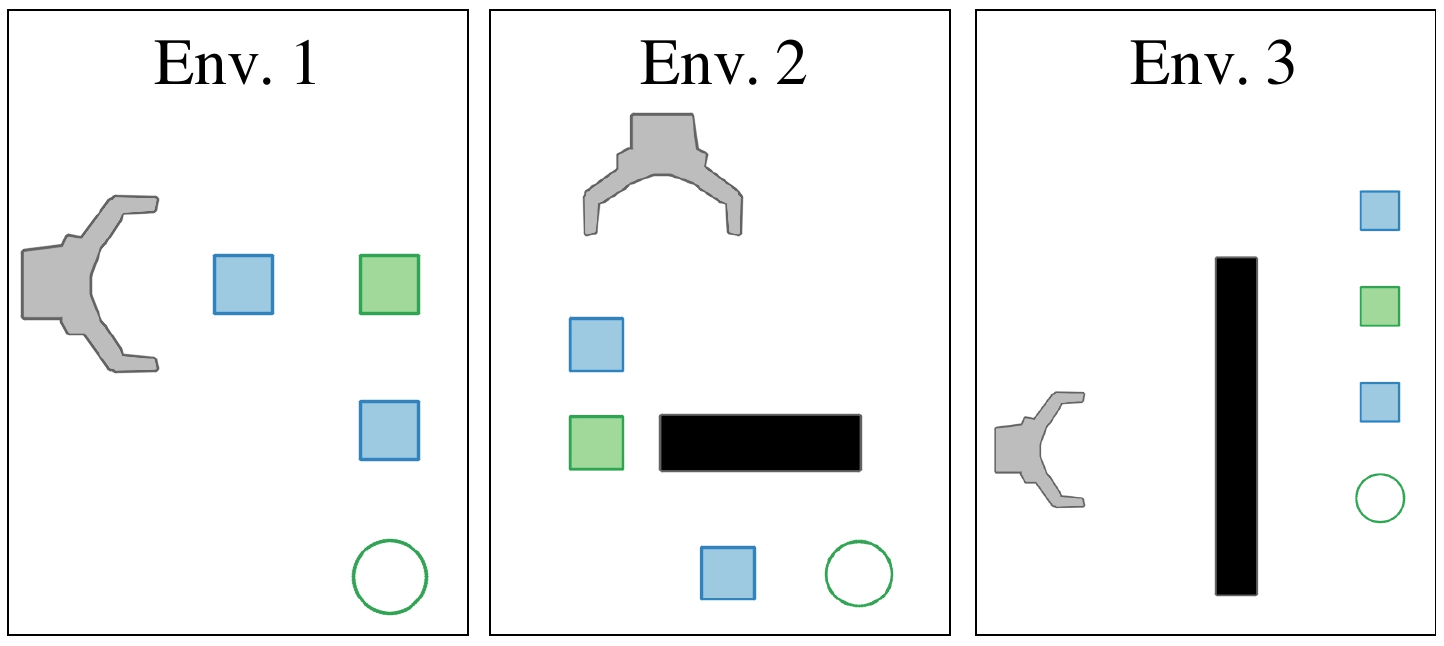}
\end{center}
\caption{The three rearrangement environments we consider. The robot (gray) must manipulate the target object (green) to the goal region (green circle) in the presence of movable objects (blue) and immovable obstacles (black).}
\label{fig:envs}
\end{figure}

To study the rearrangement manipulation problem, we consider three environments (\figref{fig:envs}). For all three, the BarrettHand ~\cite{hasan2013modelling}, which acts as the robot, has to manipulate the target object to the goal in the presence of two other movable objects. Similar to the task formulations in ~\citep{king2016rearrangement}, the final positions of the other movable objects do not matter. In Env. 2 and Env. 3, an additional immovable obstacle makes the tasks harder. To simulate quasi-static physics, needed for the $\planner$ planner, we use Box2D~\cite{catto2011box2d}. For full contact physics modelling, needed for learning, we use the MuJoCo simulator~\cite{todorov2012mujoco}. 
The planning environment has deterministic dynamics, while the physics parameters for the learning environment are sampled from a Gaussian distribution.
Since we are performing tabletop rearrangement manipulation, the C-state space for each physical entity is the SE(2) state on the plane of the table. The action applied is the SE(2) velocity of the robot on the table's plane. For more detail on these environments, we direct the reader to the Appendix.

\subsection{Speedup with Planned Episodic Resets}
Our first claim is that resetting the episode from planner states improves sample efficiency, which solves the rearrangement task faster. To demonstrate this, we plot learning curves in \figref{fig:curves} and compare the effect of planner resets with vanilla HER. Across all three environments, learning with planned episodic resets is significantly faster. Env. 2 shows the most gains, where planned episodic resets can learn policies 5X faster.  

\subsection{Effect of Model Relaxation in Planning}
Our second claim is that the model used for planning can be further relaxed while still speeding up learning with $\our$. To demonstrate this, we use the \textit{Weld} contact model, where the target is rigidly attached to the robot upon first contact. Details of the Weld model can be found in the Appendix. This simpler model speeds up planning time and hence can solve harder rearrangement tasks. Although Weld $\our$ is slower than quasi-static $\our$, we show in \figref{fig:curves} that it significantly outperforms solutions using no planner at all.

\begin{figure}[]
\begin{center}
\includegraphics[width=\linewidth]{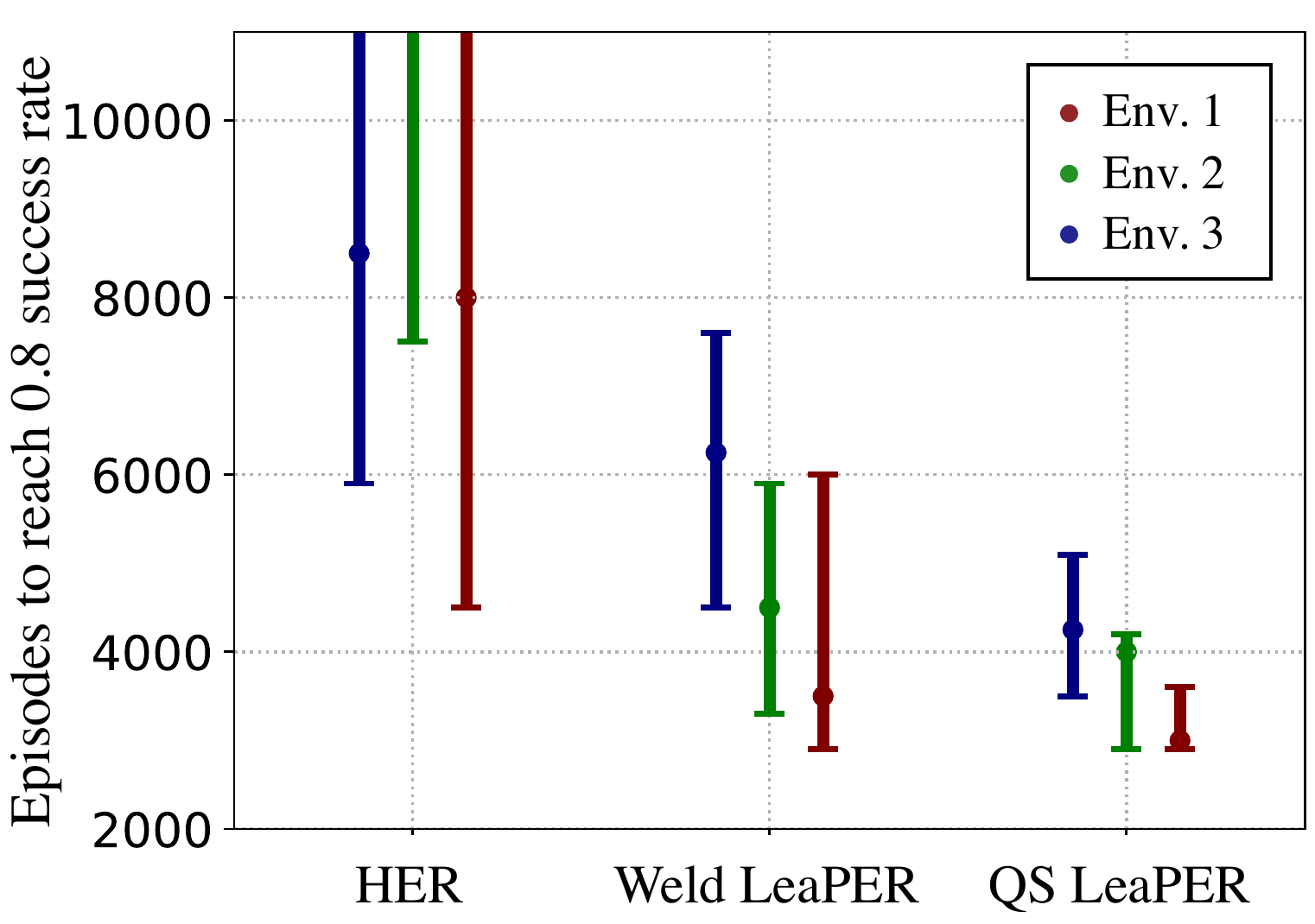}
\end{center}
\caption{The median number of episodes needed to reach an 80\% success rate across different planner models and rearrangement environments. The bars indicate the $20^{\text{th}}$ and the $80^{\text{th}}$ percentile success rate across 20 seeds.}
\label{fig:relax}
\end{figure}

To further demonstrate this effect, \figref{fig:relax} shows that different planning models affect the number of episodes $\our$ needs to solve rearrangement manipulation tasks. For all the environments we consider, Weld $\our$ not only outperforms vanilla HER but it is more robust to the initialization of the base RL algorithm.

\subsection{Comparison to Planned Trajectories}
\begin{figure*}[ht!]
\begin{center}
\includegraphics[width=7.0in]{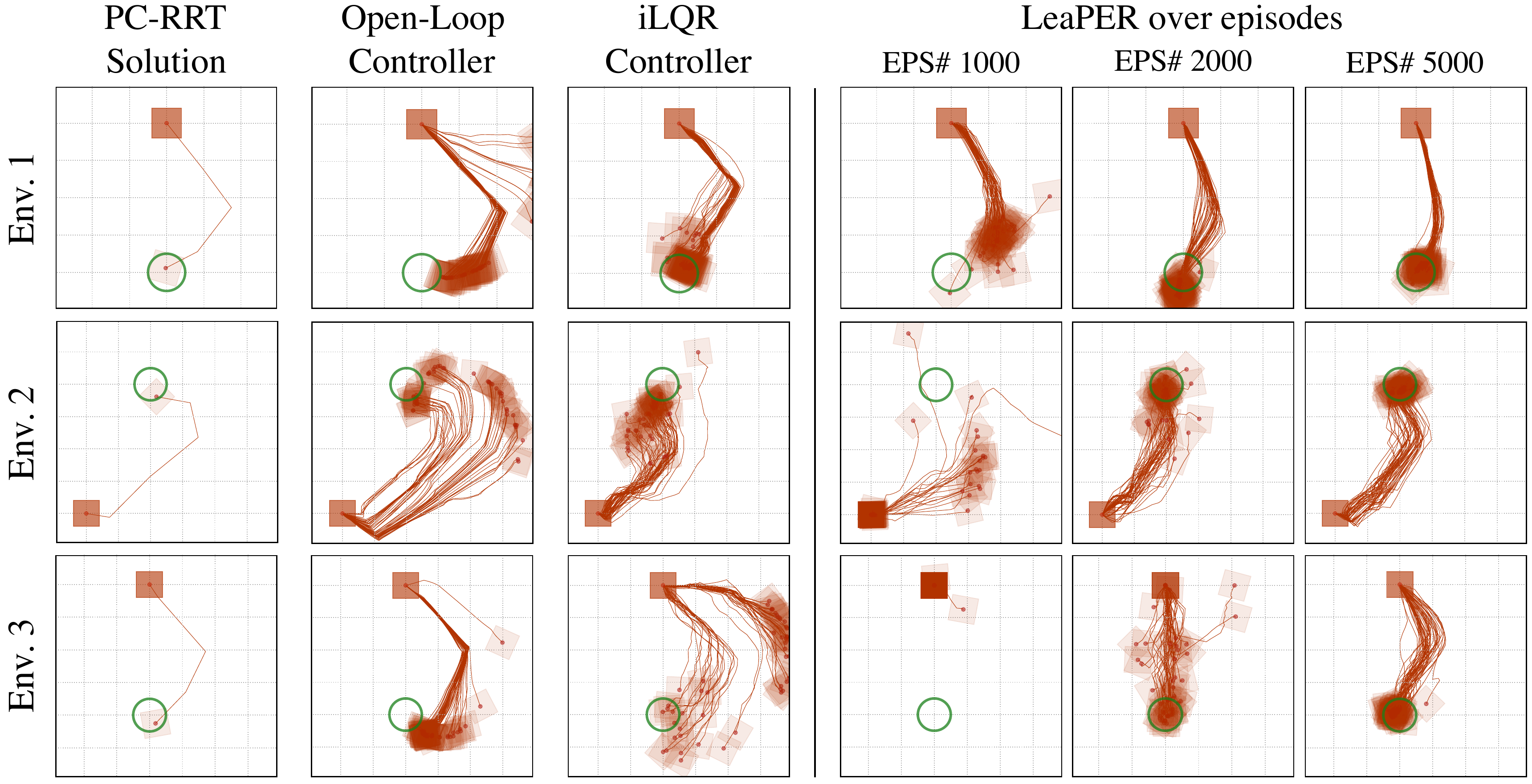}
\end{center}
\caption{On the left, trajectories of the target object are illustrated with (1) the $\planner$ solution on quasi-static dynamics, (2) a velocity controller on open-loop robot states in the full dynamics simulator, and (3) an iLQR controller on the open-loop robot states. On the right, trajectories from the policy learned after 1000, 2000, and 5000 episodes of training with $\our$. 
}
\label{fig:trajs}
\end{figure*}

Our final claim is that closed-loop policies learned by $\our$ are substantially more successful than either open-loop trajectories given by the planner or simple closed-loop controllers that track those trajectories. We compare the trajectories of the target object with different instantiations of transition dynamics in the MuJoCo simulator. \figref{fig:trajs} shows how $\our$ policies correct for deviations and reach the desired goal state. In contrast, running the open-loop trajectory with a velocity controller fails to generate optimal behaviour (\tabref{tab:trajs}). 

\begin{table}[]
\centering
\begin{tabular}{rccc}
\toprule
       & Open-Loop  & iLQR  & LeaPER (EPS\# 5000) \\ \midrule
Env. 1 & 0.00       & 0.78  & \textbf{1.00}                \\
Env. 2 & 0.06       & 0.08  & \textbf{0.92}                \\
Env. 3 & 0.02       & 0.14  & \textbf{0.99}                \\
\bottomrule
\end{tabular}
\caption{Success rates of different controllers.}
\label{tab:trajs}
\end{table}

One way to improve the simple velocity controller is to compute iLQR controllers~\cite{li2004iterative} around the desired trajectory $\xi_P$. To do this, we use a quadratic cost function and initialize the controls with the open loop control values. We note that in Env. 1, iLQR controllers produce better results than the simple velocity controller, but they are not as good as $\our$'s learned policies. For Env. 2 and Env. 3, obtaining controllers with longer time horizons is challenging and fails to significantly improve the velocity controller's performance.


\subsection{Real Robot Experiments}

To transfer the learned policies to a physical 7-DOF robot manipulator, we train the policy with randomized physics parameters similar to domain randomization in \citep{peng2017sim}. Since the inputs to the policy are the SE(2) states of the manipulator's end-effector and the three movable objects, we use an OptiTrack motion capture system to track these positions. To make the policy robust to tracking errors, we also add uniform observation noise of $\pm 0.01$ meters and $\pm 0.1$ radians during training.

Since the policy outputs SE(2) end-effector velocities every 0.1 seconds, we apply this velocity for 0.1 seconds on the robot. This end-effector velocity is realized via vector field planning~\cite{mussa1992vector}. We note that due to communication delays in fetching tracked states from the OptiTrack, feeding these states through the policy to get the action, and finally applying this action, the execution of every step exceeds 0.1 seconds. However, since we train with transition dynamics and observation randomness, the policy is able to handle these delays.
The policy is visualized in~\figref{fig:desc}. Additional examples are available in \url{https://youtu.be/feS-zFq6J1c}.


\section{Related Work}


\paragraph{Rearrangement Planning.}
Rearrangement planning can be considered a subset of manipulation planning ~\cite{latombe2012robot}, where both the motion of a robot and the objects must be considered while avoiding obstacles. Multiple works ~\cite{stilman2007navigation,nieuwenhuisen2008effective} have explored planning for a robot to reach a goal configuration. For rearrangement, however, we also need objects to reach their desired configurations. Rearrangement planning~\cite{wilfong1991motion,ota2004rearrangement,krontiris2015dealing} addresses the general form of this problem where every movable obstacle has a desired target configuration.

Early work in this field ~\cite{alami1994two,simeon2004manipulation} focused on \textit{pick-and-place} operations, where objects could be moved by grasping them. A more relevant and easier problem to solve is when only a single object must reach its goal, while the final positions of other objects do not matter. Notable work ~\cite{stilman2007manipulation,krontiris2016efficiently} solved this problem by efficiently clearing obstructions to the target object.

One of the challenges with planning to \textit{pick} objects is its limitation to objects that are light and easily graspable. Nonprehensile manipulation like \emph{pushing}~\cite{lynch1996stable} can be applied to a wider variety of objects, and its motions can be executed faster than grasping. However, the mechanics of pushing are complex and must be integrated into the planning process to solve the two-point boundary value problem (BVP). One way to solve this is to use a quasi-static contact model~\cite{whitney1982quasi} and reduce the planning problem to computing a Dubins path~\cite{zhou2017pushing}. This reduction, however, is valid only for single object manipulation with a single point of contact. To extend nonprehensile planning to multiple object interactions, physics simulators can be used to embed the physics model into the planner. \citep{king2015nonprehensile} and \citep{king2016rearrangement} demonstrate this by using a multi-body, quasi-static contact model in Box2D with a kinodynamic RRT planner to push multiple objects with a robot. The feasibility of these plans depends on how closely real-world physics matches quasi-static physics. In practice, due to errors in system identification and environment modelling as well as non quasi-static interactions, open-loop controls can deviate from the planned trajectory, and objects can fail to reach their goal.

\paragraph{Learning for Rearrangement Manipulation.}
Model-free reinforcement learning has achieved considerable success in playing Atari games~\cite{mnih2015human}, simulated locomotion~\cite{schulman2015trust,lillicrap2015continuous} and manipulation~\cite{levine2016end}. However, transferring these algorithms to physical robots has been challenging due to the poor sample complexity of model-free learning and the \textit{reality gap} between simulators and the real world. Randomizing physics and visual observations has shown promise in overcoming this gap~\cite{sadeghi2016cad,peng2017sim,pinto2018asymmetric}. To alleviate poor sample complexity, several works have investigated large-scale data collection for simple tasks like grasping~\cite{pinto2016supersizing,levine2016learning,gupta2018robot} and pushing~\cite{agrawal2016learning,finn2017deep,pinto2016curious}. However, rearrangement manipulation is a long horizon and sparse reward task that involves manipulating multiple objects, rendering pure model-free learning undesirable. 

One way to improve the sample complexity of model-free learning is to incorporate model-based information~\cite{sutton1991dyna} during learning. \citep{deisenroth2011pilco} learn the dynamics model using Gaussian Processes and use rollouts with this model for learning closed-loop policies. \citep{nagabandi2017neural} use learned dynamics models as a prior for model-free learning, while \citep{bansal2017mbmf} use model-based priors for learning. Although these outperform pure model-free methods, they do not exploit the structure of these models. One way to do so is to plan to get an open-loop trajectory that is then executed with a learned low-level policy. However, this assumes that we have exact models for planning, and hence it has only been helpful for simple navigation tasks~\cite{faust2017prm}. Instead of being constrained by the planner's solutions or an approximate transition model, $\our$ uses the planner's solution as a prior for learning.

\section{Discussion}
In this work, we presented a sample-efficient yet simple method for learning nonprehensile rearrangement policies. We do this by resetting episodes with the solution from rearrangement planning during learning. The closed-loop policies learned from our method, $\our$, can successfully perform several rearrangement tasks with stochastic dynamics. Furthermore, we can learn these policies even with extremely relaxed planning dynamics like the Weld contact model. 

Although we use a powerful method, HER, as our base RL algorithm, we note that our method is agnostic to the type of base RL algorithm. However,it requires an episodic environment where resets are possible. Since we use the MuJoCo environment to learn, resetting the environment the desired states is trivial. However, transferring $\our$ to the real world will require engineering the environment to make it resettable. But policies learned in a powerful simulator can be transferred to real robots either by performing accurate state estimation or by employing state-of-the-art \textit{Sim2Real} methods. We transfer the learned policy using OptiTrack to achieve accurate state estimation and dynamics domain randomization for the \textit{Sim2Real} transfer.

$\our$ is not limited to the domain of rearrangement manipulation: planning has been used in several other domains where controllers are frequently designed with simplified models in mind. One example of this is walking, where the spring-loaded inverted pendulum~\cite{schwind1998spring} and the zero-moment point~\cite{vukobratovic2004zero} models are used to generate fast plans. We believe that $\our$ can generate robust closed-loop solutions in these domains as well.







\section{Acknowledgments}
We thank Gilwoo Lee for assistance with the physical robot experiments. We also thank Sandy Kaplan, Xingyu Lin and Christopher Atkeson for feedback and suggestions.
This work was supported by a NASA Space Technology Research Fellowship (\#80NSSC17K0137), the National Institute of Health R01 (\#R01EB019335), National Science Foundation CPS (\#1544797), National Science Foundation NRI (\#1637748), the Office of Naval Research, the RCTA, Amazon, and Honda.

\fontsize{9.2pt}{10.2pt} \selectfont 
\bibliography{references}
\bibliographystyle{aaai}
\clearpage
\section{Appendix}


\subsection{Environment Details}
\label{app:envs}
All the movable objects, including the target object, are cubes with a side-length of 8cm. The obstacle in Env. 2 is 30~cm long and 8~cm wide, while the obstacle in Env. 3 is 70~cm long and 8~cm wide. Each time-step in the MuJoCo simulation is 0.1 seconds, so the policy acts at 10 Hz. The episode length for Env. 1 and Env. 2 is 50 timesteps, and for Env. 3 is 150 timesteps. Contacts are enabled for all the physical entities in the workspace (table, target object, movable objects, obstacles, and robot). We use regular frictional contacts, which can generate both normal force and tangential friction force opposing slip (equivalent to using $\textit{condim}=3$ in MuJoCo).

The action is the robot's SE(2) velocity, which is 3-dimensional. For stable simulations, the action velocity $(\dot{x}, \dot{y}, \dot{\theta})$ of the robot is limited to (0.25~m/s, 0.25~m/s, 2.5~rad/s). The observation from the environment is the $(x,y,\textrm{sin}(\theta),\textrm{cos}(\theta))$ of every movable entity in a global frame, so the observation space is 16-dimensional. The goal region is a disc of radius 5~cm. If the target object does not come within 5~cm of the desired goal, it gets a reward of $-1$.

Before every episode, a random physics parameter is sampled from a normal parameter distribution and held constant throughout the episode. These parameters include the mass of movable entities and the friction of all physical entities. The mean is the nominal parameter value, and the standard deviation is twice the magnitude of the nominal parameter. If a sampled parameter is invalid, it is re-sampled. For training policies to run on the robot, we add an observation noise of 1~cm to the positions and 0.1~rad to the angles of every movable entity.


\subsection{Training Details}

We build on the OpenAI Baselines framework~\cite{hesse2017openai} to train our policies with HER. Our actor and critic networks both contain 3 hidden layers with 256 neurons each. The Adam~\cite{kingma2014adam} optimizer is used with an initial learning rate of 0.01. The number of goals set for replay in HER is 4. While applying actions, a random action is chosen 30\% of the time, and a Gaussian with standard deviation of 0.2 is applied on the predicted action for the remaining 70\%. These parameters are similar to HER defaults and we did not need to tune them.  


\subsection{The Quasi-Static Pushing Model}

\begin{figure}[h]
\begin{center}
\includegraphics[width=3.2in]{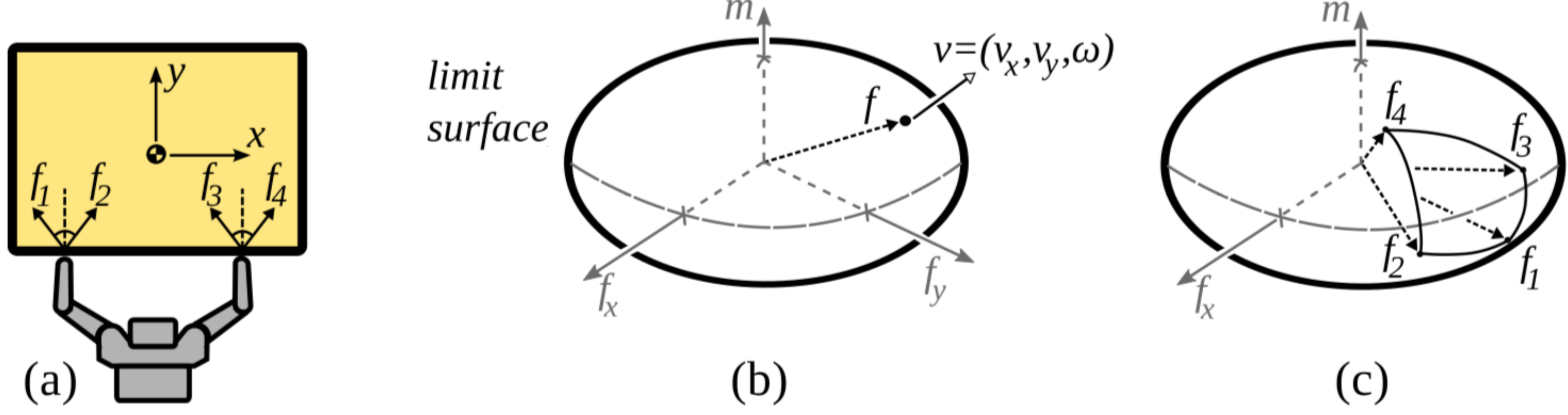}
\end{center}
\caption{The quasi-static pushing model described in \cite{king2013pregrasp}. (a) Friction cones at the contacts. (b) The limit surface. (c) The cone of all possible forces that the hand can apply.}
\label{fig:app_quasi}
\end{figure}

Consider the robot hand and object in \figref{fig:app_quasi}a, where the friction cones~\cite{mason1986mechanics} can be computed from the contact friction coefficient. The limit surface~\cite{goyal1991planar} relates the generalized forces applied on the object to the resulting generalized velocity. With finite pressure, the limit surface is similar to the three-dimensional ellipsoid in \figref{fig:app_quasi}b. Here, the dimensions are the force in the x-direction $f_x$, the force in the y-direction $f_y$, and the moment $m$. 

In quasi-static pushing, the generalized forces are constrained to the limit surface, and the resulting generalized velocity $(v_x, v_y, \omega)$ can be computed using the normal to the surface. \figref{fig:app_quasi} shows the friction cone and three-dimensional cone corresponding to the applied force. With quasi-static physics, given an applied force and friction coefficient, the object is constrained to move with velocities prescribed by the segment on the limit surface. 

\begin{figure*}[!b]
\centering
  \begin{subfigure}[b]{0.3\textwidth}
	\centering
    \includegraphics[height=15em]{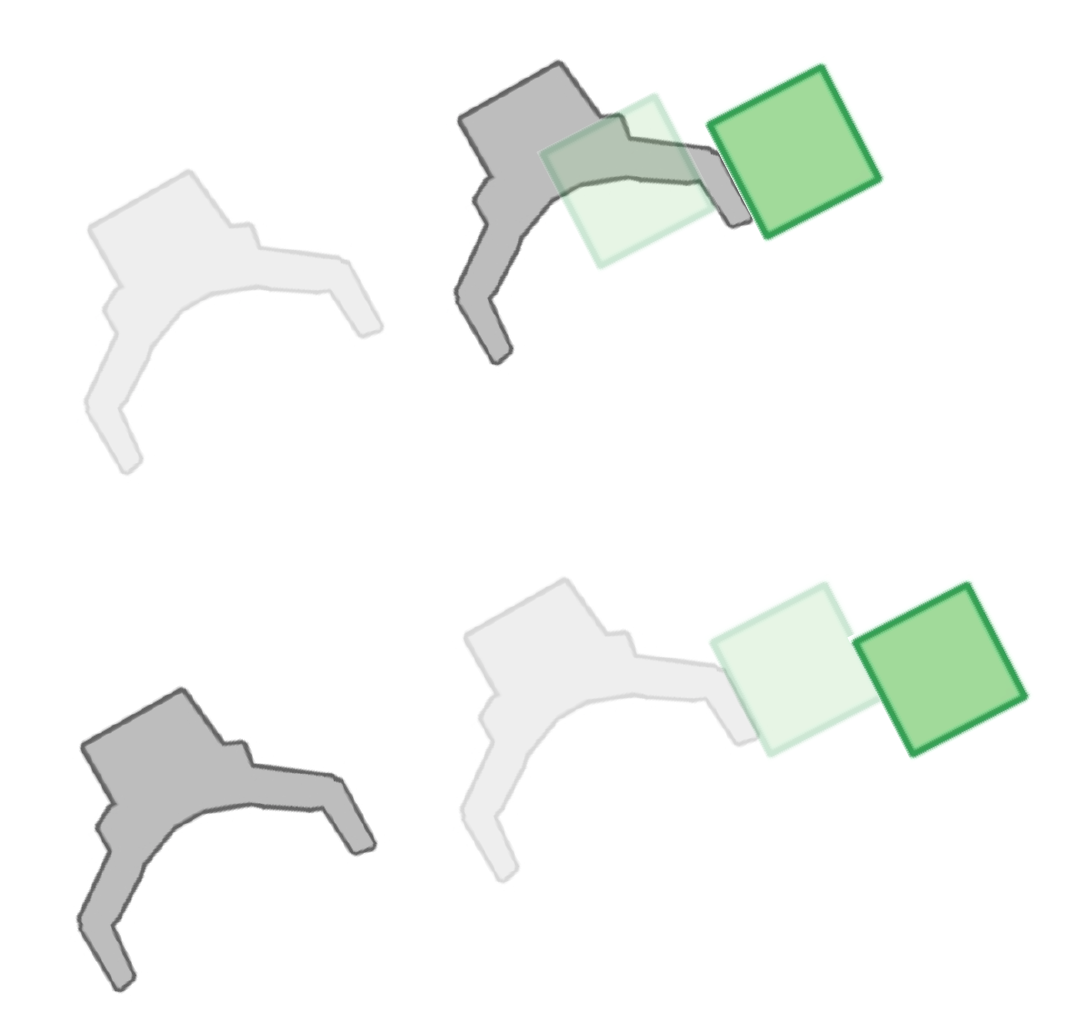}
    \caption{Dynamic model}
    \label{fig:dynamic}		
  \end{subfigure}
  \begin{subfigure}[b]{0.3\textwidth}
	\centering
	\includegraphics[height=15em]{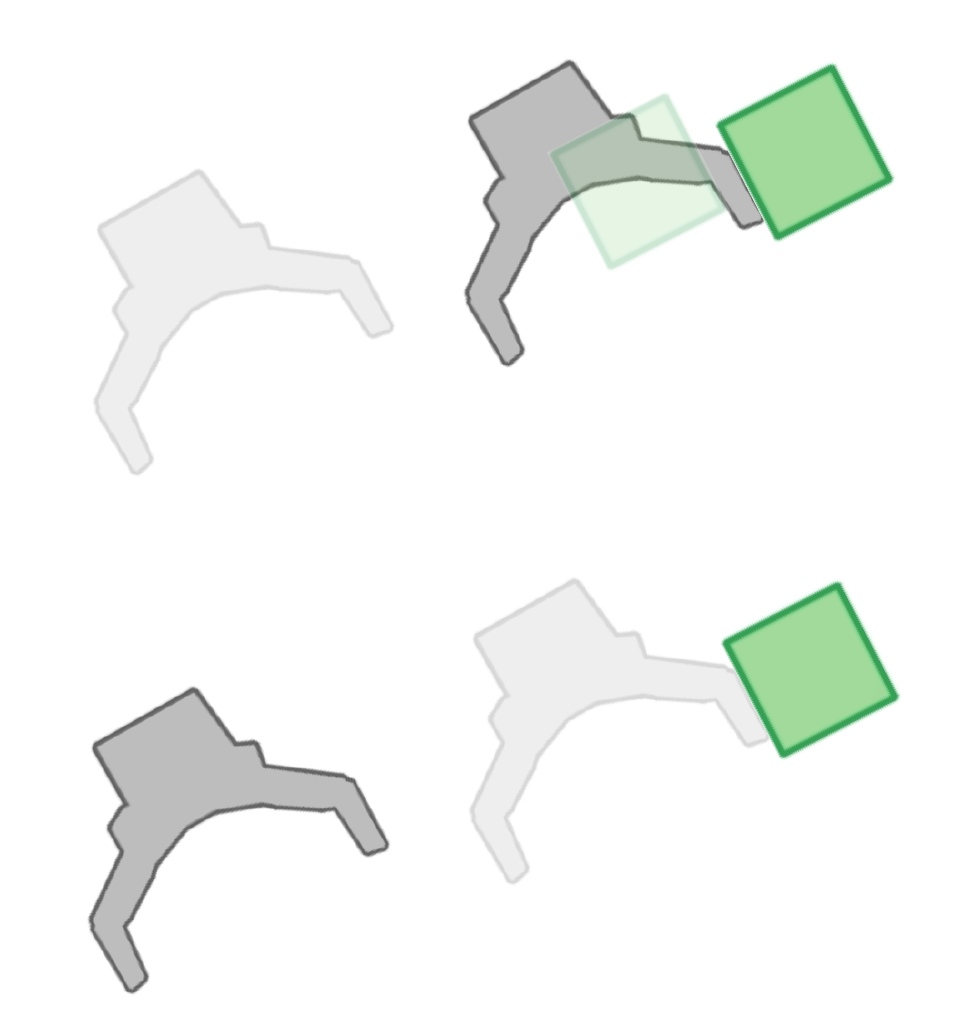}
	\caption{Quasi-static model}
	\label{fig:quasistatic}
  \end{subfigure}
  \begin{subfigure}[b]{0.3\textwidth}
	\centering
	\includegraphics[height=15em]{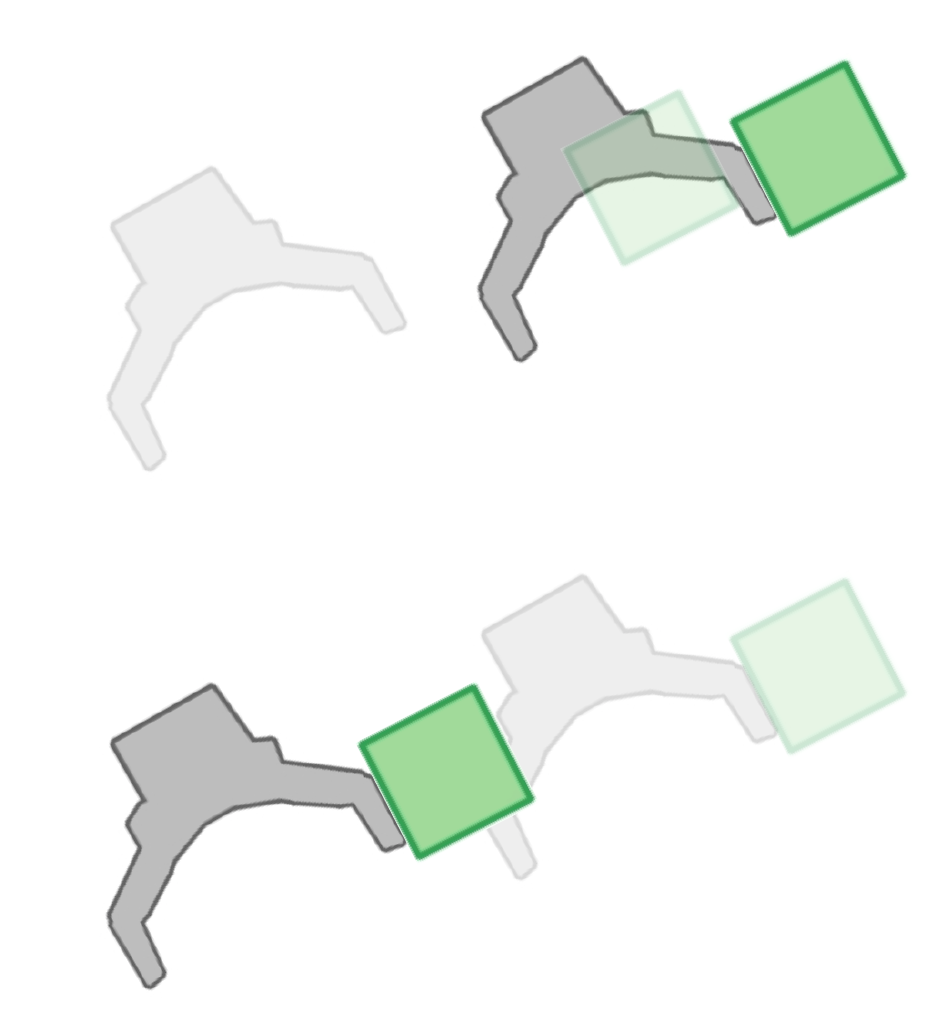}
	\caption{Weld model}
	\label{fig:weld}
  \end{subfigure}
  
\caption{We visualize three physics models (Dynamic, Quasi-static, Weld) through the same interaction. The end-effector pushes the object (top row) before moving back to its initial position (bottom row). In the Dynamic model, the object continues moving forward before stopping due to friction. In the Quasi-static model, the object comes to rest as soon as it loses contact with the end-effector. In the Weld model, the object remains rigidly attached to the end-effector.}
\label{fig:models}
\end{figure*}

\subsection{Planning Details}

We use the planning framework detailed in \cite{king2015nonprehensile}, which extends the Open Motion Planning Library (OMPL) \cite{ompl}. We utilize Box2D \cite{catto2011box2d} to simulate the physics interactions in 2D since we constrain and plan for the end-effector on a plane parallel to the table. We enforce the quasi-static~(\figref{fig:quasistatic}) and relaxed Weld~(\figref{fig:weld}) physics models in the simulator. As we expect, the planning times are significantly lower with the relaxed physics model. For Env. 3, the average planning time with the quasi-static model is 35.2 seconds across 10 trials. With the Weld contact model, we obtain a 2X speed-up, with an average planning time of 16.1 seconds.

\begin{figure*}[h]
\centering
\includegraphics[width=7.0in]{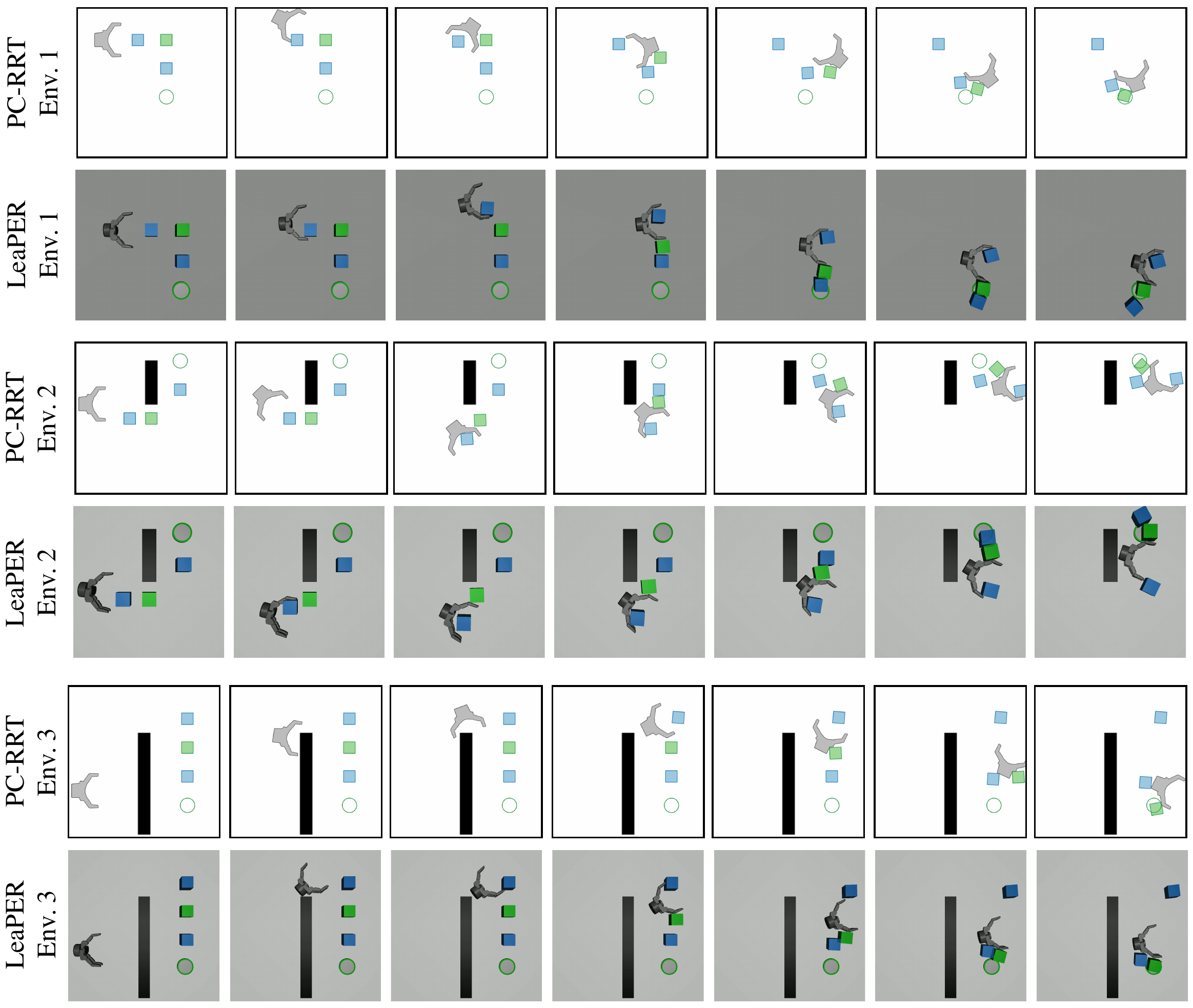}
\caption{$\planner$ solutions with quasi-static dynamics and $\our$ policy with full contact dynamics on the three rearrangement environments.}
\label{fig:app_sol}
\end{figure*}

\end{document}